%% file: main.tex
\documentclass{article}

\usepackage[numbers]{natbib}
\usepackage[preprint]{neurips_2023}
\usepackage[dvipsnames]{xcolor}         
\usepackage[utf8]{inputenc} 
\usepackage[T1]{fontenc}    
\usepackage[colorlinks=true,linkcolor=linkColor,citecolor=linkColor,filecolor=linkColor,urlcolor=linkColor]{hyperref}       
\usepackage{url}            
\usepackage{booktabs}       
\usepackage{amsfonts}       
\usepackage{nicefrac}       
\usepackage{microtype}      
\usepackage{graphicx}
\usepackage{algorithm}
\usepackage{algpseudocode} 
\usepackage{ragged2e}
\usepackage{amsmath}
\usepackage{array}          
\usepackage{tabularx}
\usepackage{colortbl}
\usepackage{float}
\usepackage{comment}

\definecolor{linkColor}{rgb}{0.18,0.39,0.62}

\input{settings.tex}
\input{math_commands.tex}

\title{Meta Reasoning for Large Language Models}

\author{
  Peizhong Gao\textsuperscript{1}\thanks{Work was done when interning at Microsoft Research Asia.} ~~ Ao Xie\textsuperscript{1} ~~ Shaoguang Mao\textsuperscript{2} ~~ Wenshan Wu\textsuperscript{2} \\
  \textbf{Yan Xia}\textsuperscript{2} ~~ \textbf{Haipeng Mi}\textsuperscript{1} ~~ \textbf{Furu Wei}\textsuperscript{2}\\
  \textsuperscript{1}\space\texttt{Tsinghua University} \\
  \textsuperscript{2}\space\texttt{Microsoft Research} \\
  \href{https://aka.ms/GeneralAI}{https://aka.ms/GeneralAI}
}

\begin{document}

\maketitle
\vspace{-0.7cm}
\begin{abstract}
\vspace{-0.25cm}
We introduce Meta-Reasoning Prompting (MRP), a novel and efficient system prompting method for large language models (LLMs) inspired by human meta-reasoning. Traditional in-context learning-based reasoning techniques, such as Tree-of-Thoughts, show promise but lack consistent state-of-the-art performance across diverse tasks due to their specialized nature. MRP addresses this limitation by guiding LLMs to dynamically select and apply different reasoning methods based on the specific requirements of each task, optimizing both performance and computational efficiency. With MRP, LLM reasoning operates in two phases. Initially, the LLM identifies the most appropriate reasoning method using task input cues and objective descriptions of available methods. Subsequently, it applies the chosen method to complete the task. This dynamic strategy mirrors human meta-reasoning, allowing the model to excel in a wide range of problem domains. We evaluate the effectiveness of MRP through comprehensive benchmarks. The results demonstrate that MRP achieves or approaches state-of-the-art performance across diverse tasks. MRP represents a significant advancement in enabling LLMs to identify cognitive challenges across problems and leverage benefits across different reasoning approaches, enhancing their ability to handle diverse and complex problem domains efficiently. Every LLM deserves a Meta-Reasoning Prompting to unlock its full potential and ensure adaptability in an ever-evolving landscape of challenges and applications.
\end{abstract}

\begin{figure}[ht]
    \centering
    \includegraphics[width=0.8\textwidth]{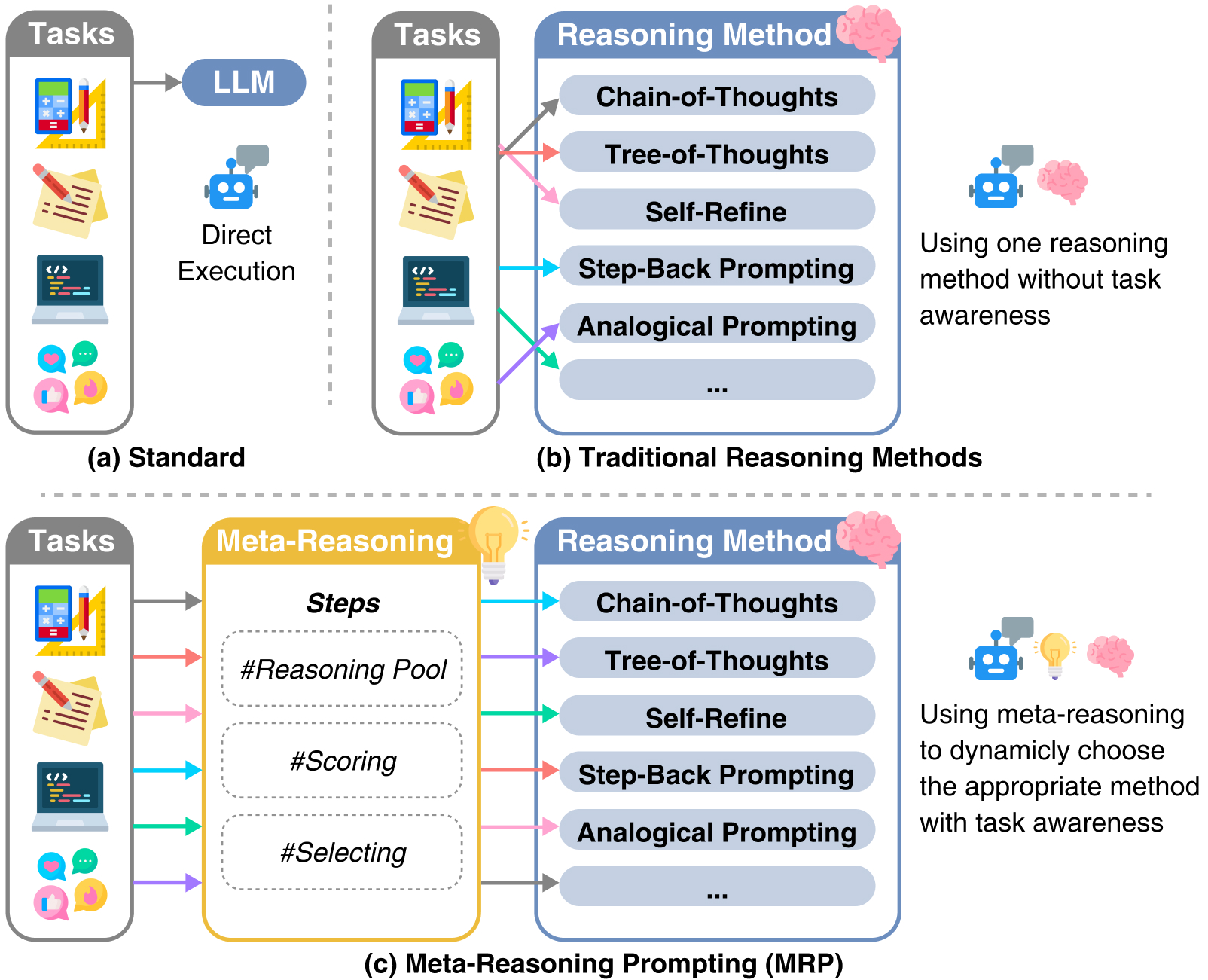}
    \caption{Illustration of Meta-Reasoning Prompting (MRP) and the difference compared to standard reasoning and traditional reasoning methods.}
    \label{MRP framework}
\end{figure}

\section{Introduction}
Large language models (LLMs) have shown remarkable capabilities in natural language understanding and generation, making significant strides in various reasoning tasks. However, the diversity and complexity of real-world problems require advanced reasoning methods that surpass the capabilities of a single, static approach. While existing reasoning techniques, such as Chain-of-Thoughts~\cite{COT}, Tree-of-Thoughts~\cite{TOT}, Analogical Prompting~\cite{yasunaga2023large}, and Solo Performance Prompting~\cite{SPP}, offer valuable tools for enhancing reasoning, they often fall short in consistently achieving state-of-the-art performance across different tasks.

These challenges highlight the need for a more adaptive and flexible approach to reasoning in LLMs. In human cognition, meta-reasoning involves monitoring and regulating reasoning and problem-solving activities, adjusting strategies based on the context and specific task requirements~\cite{cox2011metareasoningbook,cox2011metareasoning}. This adaptive capability allows humans to efficiently allocate cognitive resources, balancing trade-offs between accuracy, complexity, and computational cost. Inspired by this, we propose Meta-Reasoning Prompting (MRP) to endow LLMs with similar adaptive reasoning capabilities.

Meta-Reasoning Prompting (MRP) is a simple yet effective system prompt designed to guide LLMs in dynamically selecting and applying the most suitable reasoning method for a specific task. By incorporating meta-reasoning principles, MRP transforms task-specific prompt engineering into a more general and flexible approach. Under the guidance of MRP, the LLM evaluates the task input and selects an appropriate reasoning method from a predefined set (Reasoning Pool). This selection is informed by objective descriptions and evaluations of the available methods. The chosen method is then applied to complete the task, ensuring the model uses the most effective strategy for the given problem.

Recent advances in reasoning techniques, such as those described in \cite{yang2024buffer, wang2024mixture}, introduce a meta-buffer for storing high-level thoughts or use ensemble mechanisms to improve model generalizability. While some of these approaches align with the inherent logic of meta-reasoning, our proposed MRP achieves simple and efficient meta-cognitive effects by directly leveraging the meta-reasoning capabilities of LLMs through prompts, without introducing complex mechanisms.

To evaluate the effectiveness of MRP, we conducted experiments using multiple widely used benchmarks. These benchmarks cover different knowledge and reasoning abilities, providing a comprehensive test of the LLM's performance across various reasoning tasks. Our findings demonstrate that MRP not only approaches state-of-the-art performance across these benchmarks but also excels in tasks requiring a blend of different reasoning strategies. Additionally, we observe that larger models, such as GPT-4, exhibit superior meta-reasoning capabilities compared to smaller models like GPT-3.5.

As models improve, their understanding of problems and methods—i.e., their meta-reasoning abilities—also enhances. MRP utilizes the inherent meta-cognitive abilities of LLMs, providing a straightforward and effective method that enhances their generality across different tasks. Experimental and analytical results indicate the significant potential of MRP in boosting LLM performance. Future work could explore the broader application of MRP, such as constructing training data to enhance the meta-cognitive and general reasoning abilities of LLMs during the training process.

Our key contributions are as follows:

\begin{enumerate}
    \item We propose Meta-Reasoning Prompting (MRP), a system prompt that enables LLMs to dynamically select the most suitable reasoning method for specific tasks, enhancing their flexibility and effectiveness.
    \item Experiments on multiple benchmarks show that MRP approaches state-of-the-art performance and excels in tasks requiring diverse reasoning strategies, particularly in larger models like GPT-4.
    \item MRP leverages LLMs' inherent meta-cognitive abilities, improving their generality and performance across tasks. Future work could further enhance these abilities through targeted training data.
\end{enumerate}

\begin{figure}[t]
    \centering
    \includegraphics[width=0.9\columnwidth]{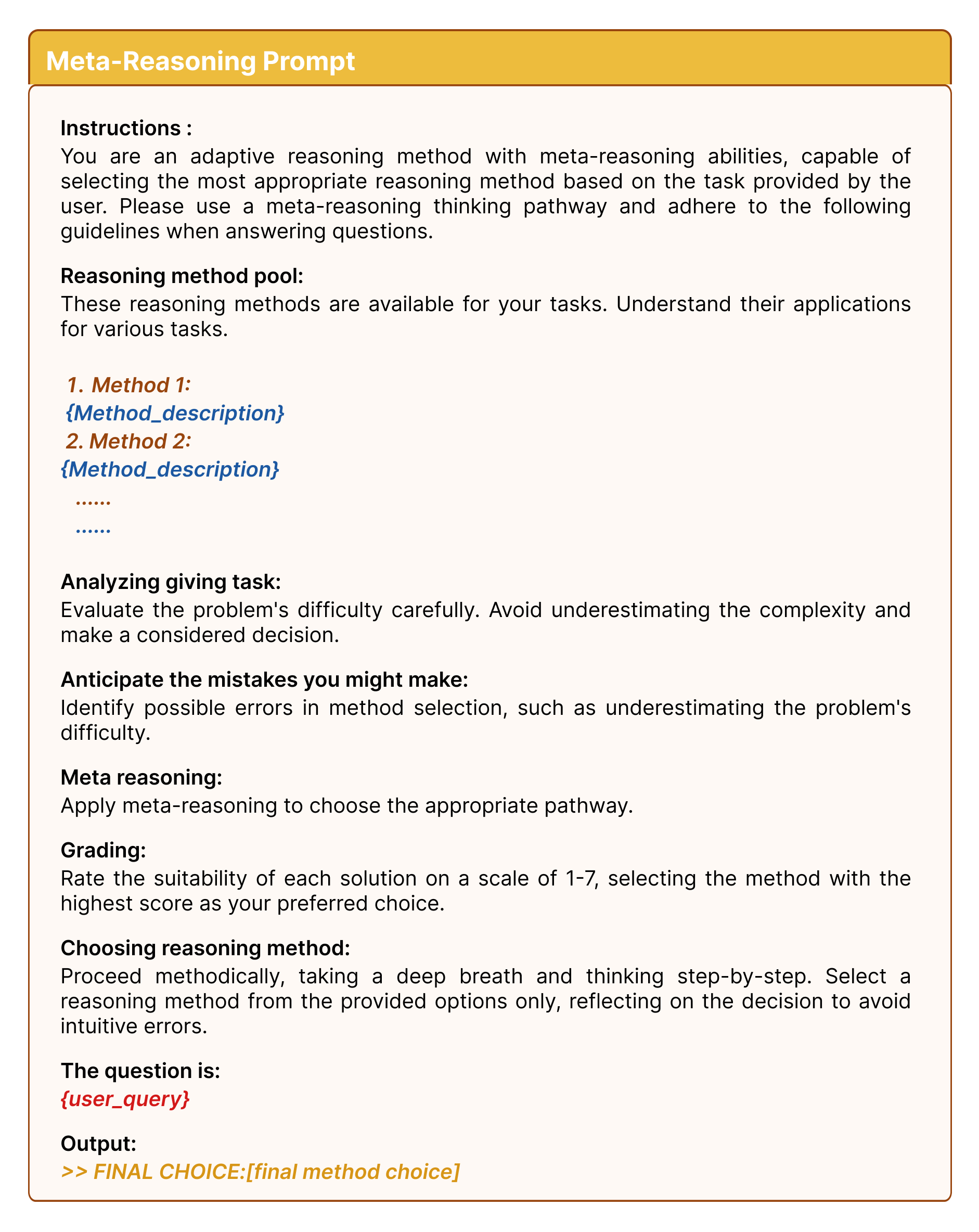}
    \vspace{-0.5cm}
    \caption{Meta-Reasoning Prompt.}
    \label{Fig:MRPprompt}
    \vspace{-0.5cm}
\end{figure}

\section{Meta-Reasoning Prompting}
The Meta-Reasoning Prompting (MRP) is designed to guide a Language Learning Model (LLM) in selecting the most suitable reasoning method from a pool of available methods, thereby enhancing the overall reasoning performance of the model. Detailed prompts can be found in Fig. \ref{Fig:MRPprompt}.

Guided by the Meta-Reasoning Prompting, the LLM ($M$) begins with an input $x_0$ and a set of available reasoning methods ${\alpha_1, \alpha_2, \ldots, \alpha_n}$. A reasoning pool contains descriptions of each reasoning method in the form of prompts ${p_1, p_2, \ldots, p_n}$, with these descriptions extracted from the abstracts of corresponding papers. A Meta-Reasoning Prompting $p_{MR}$ is defined to guide the selection process. For each reasoning method $\alpha_i$ ($i$ ranging from 1 to $n$), the model $M$ evaluates the combined prompt $(p_i | p_{MR} | x_0)$. This evaluation yields a score $s_i$ indicating the effectiveness of method $\alpha_i$ for the given input $x_0$.

\begin{equation}
  s_i = M(p_i \| p_{MR} \| x_0) \quad \text{for} \quad i = 1, 2, \ldots, n.  
\end{equation}

The algorithm identifies the reasoning method $\alpha_k$ that receives the highest score $s_i$ by finding the index $k$ that maximizes the set ${s_1, s_2, \ldots, s_n}$.

\begin{equation}
    k = \arg\max_{i} \{s_1, s_2, \ldots, s_n\}
\end{equation}

Once the best reasoning method $\alpha_k$ is determined, it is executed on the input $x_0$. The model $M$ generates the final output $y_0$ using the prompt $(p_k | x_0)$, which combines the description of the chosen reasoning method with the original input.
\begin{equation}
    y_0 = \alpha_k(x_0)
\end{equation}

\begin{algorithm}
\caption{Process Under MRP Guidelines}
\textbf{Require:} Input $x_0$, model $M$, reasoning methods $\alpha_1, \alpha_2, \ldots, \alpha_n$, a set of prompts of reasoning method descriptions $p_1, p_2, \ldots, p_n$, the prompt for meta reasoning process denoted as $p_{MR}$. Evaluation score is denoted as $s_1, s_2, \ldots, s_n.$
\begin{algorithmic}[1]
\For{$i = 1$ \textbf{to} $n$}
    \State $s_i = M(p_i \| p_{MR} \| x_0)$
\EndFor
\State $k = \arg\max_{i} \{s_1, s_2, \ldots, s_n\} $
\State Determine $k$ for which $\alpha_k$ is executed and reason with the chosen method.
\State $y_0 = \alpha_{k}(x_0)$
\end{algorithmic}
\textbf{Return} $y_0$
\end{algorithm}

\section{Experiments}

\subsection{Setup}

\paragraph{Implementation of Meta-Reasoning Prompting} We implement MRP with seven popular and distinct in-context learning reasoning methods, which also serve as our baseline for comparison. We prompt descriptions for each method, allowing the LLM to understand.

\paragraph{Tasks}
We experiment with seven diverse tasks, Details about the dataset and its construction are provided in Appendix~\ref{sec:appendix Dataset Details}:
\begin{enumerate}
    \item \textbf{Arithmetic Reasoning:} GSM8K~\cite{GSM8K}, 1319 basic math questions. 
    \item \textbf{Complex Mathematical Reasoning:} Game of 24~\cite{TOT}, a game using 4 numbers and basic arithmetic four operations to obtain 24.
    \item \textbf{Creative Writing:} Trivia Creative Writing (Trivia CW)~\cite{SPP,triviaqa}, necessitating the model to assimilate and combine heterogeneous information from multiple domains internally.
    \item \textbf{Multi-Hop Reasoning:} HotpotQA,~\cite{hotpotqa}, requiring models to connect pieces of information from multiple documents to answer a question.
    \item \textbf{Social Reasoning:} BigToM~\cite{bigtom}, to evaluate social situations understanding and the theory of mind. 
    \item \textbf{Computer Code:} Code Readability (Code)~\cite{codereadability}, to enhance the readability of given code snippets. 
    \item \textbf{STEM:} MMLU~\cite{mmlu}, Physics, Chemistry, Biology, and Math problems of high school domain.
\end{enumerate}   
\paragraph{Metrics} To prevent any method from skewing the results due to exceptional performance on a specific task, we reported both the arithmetic mean accuracy and the harmonic mean accuracy of each method across all benchmarks.


\paragraph{Models}
We used gpt-3.5-turbo\footnote{Azure OpenAI, Model Name: gpt-35-turbo, API Version: 0301} and gpt-4-turbo\footnote{Azure OpenAI, Model Name: gpt-4, API Version: 1106-Preview} with identical prompts to compare the effect of model size on meta-reasoning ability.

\paragraph{Baselines}
We select seven popular reasoning methods as baselines. These methods include: 
\begin{enumerate}
    \item \textbf{Chain-of-Thoughts:} breaking down problems into a series of coherent reasoning steps~\cite{COT}.
    \item \textbf{Tree-of-Thoughts:} exploring multiple reasoning paths and self-evaluating choices to solve complex problems~\cite{TOT}.
    \item \textbf{Analogical prompting:} self-generating few-shots based on past experiences and related problems~\cite{yasunaga2023large}.
    \item \textbf{Self-Refine:} self-evaluating for refinement and continuously improving the output~\cite{selfrefine}.
    \item \textbf{Solo Performance Prompting:} simulating multiple personas to collaboratively solve complex tasks~\cite{SPP}. 
    \item \textbf{Step-Back Prompting:} abstract high-level concepts and principles to guide the reasoning process~\cite{TSB}.
    \item \textbf{SimToM:} enabling perspective-taking to understand the character's beliefs and goals~\cite{simtom}
\end{enumerate}

\begin{table*}
\centering
\caption{Experiments with GPT4: Comparison of performance on benchmarks using Meta-Reasoning Prompting versus using other methods independently. \textbf{Bold} represents the best performance, and \underline{underline} represents the second-best performance.}
\label{tab:benchmarking_gpt4}
{\small  
\begin{tabularx}{\linewidth}{@{}l*{9}{>{\centering\arraybackslash}X}@{}}
\toprule
\textbf{Method} & \textbf{GSM8K} & \textbf{Gameof24} & \textbf{Trivia~CW} & \textbf{HotpotQA} & \textbf{BigToM} & \textbf{Code} & \textbf{MMLU} & \textbf{Macro~Avg.} \\
\midrule
COT                       & 0.914 & 0.050 & 0.762 & \underline{0.800} & 0.470 & 0.685 & \underline{0.894} & \cellcolor{gray!10}0.654 \\
TOT                       & \textbf{0.942} & \textbf{0.410} & 0.786 & 0.716 & 0.430 & 0.765 & 0.815 & \cellcolor{gray!10} \underline{0.725} \\
Analogical           & 0.924 & 0.040 & 0.735 & 0.777 & 0.500 & 0.614 & \textbf{0.947} & \cellcolor{gray!10}0.648 \\
Self-Refine          & 0.929 & 0.080 & 0.764 & 0.763 & 0.470 & \textbf{0.872} & 0.861 & \cellcolor{gray!10}0.677 \\
SPP                       & 0.929 & 0.170 & \textbf{0.861} & 0.763 & 0.550 & 0.672 & 0.874 & \cellcolor{gray!10}0.688 \\
STEP-BACK          & 0.933 & 0.090 & 0.787 & \textbf{0.810} & 0.420 & 0.809 & 0.841 & \cellcolor{gray!10}0.670 \\
SimTom                & \underline{0.938} & 0.040 & 0.739 & 0.667 & \textbf{0.590} & 0.694 & 0.815 & \cellcolor{gray!10}0.640 \\
\textbf{MRP (our)}            & 0.921 & \underline{0.310} & \underline{0.796} & 0.797 & \underline{0.570} & \underline{0.867} & 0.854 & \cellcolor{gray!10} \textbf{0.772} \\
\bottomrule
\end{tabularx}
}
\end{table*}

\begin{figure*}[t]
    \centering    \includegraphics[width=\textwidth]{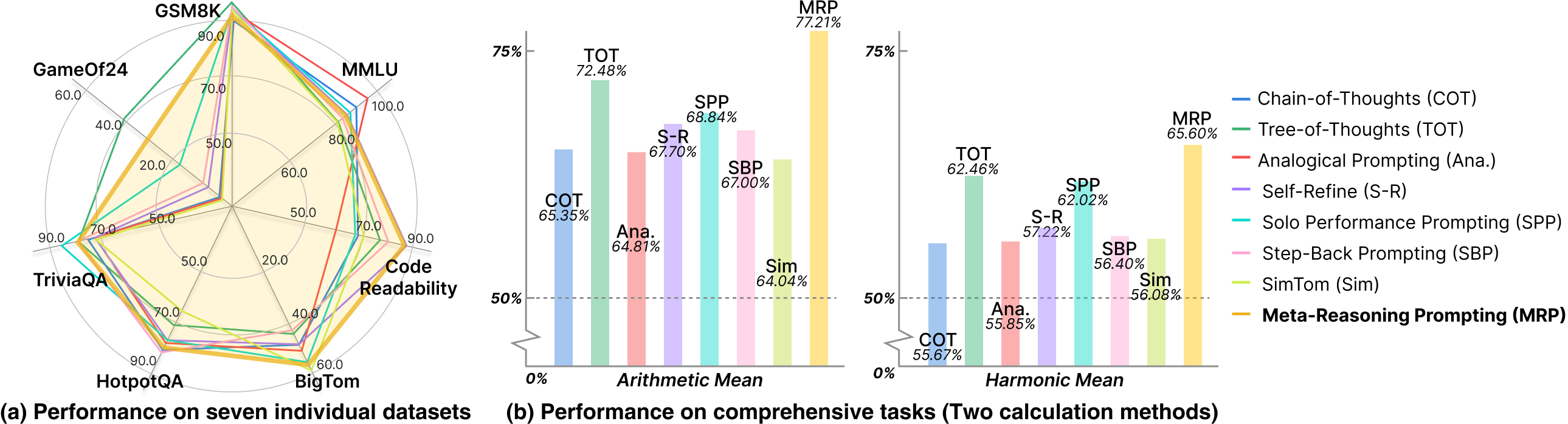}
    \caption{(a) Comparison of methods on different benchmarks reveals that guiding LLM to dynamically choose the appropriate reasoning method enables MRP to achieve consistently better performance across all tasks. (b) The arithmetic and harmonic average performances of applying a specific reasoning approach to all benchmarks demonstrate that MRP consistently excels in overall evaluation.}
    \label{fig:ladar}
\end{figure*}

\begin{table*}[htbp]
\centering
\caption{Experiments with GPT3.5: Comparison of performance on benchmarks using Meta-Reasoning Prompting versus using other methods independently. \textbf{Bold} represents the best performance, and \underline{underline} represents the second-best performance.}
\label{tab:benchmarking_gpt35}
{\small  
\begin{tabularx}{\linewidth}{@{}l*{9}{>{\centering\arraybackslash}X}@{}}
\toprule
\textbf{Method} & \textbf{GSM8K} & \textbf{Gameof24} & \textbf{Trivia~CW} & \textbf{HotpotQA} & \textbf{BigToM} & \textbf{Code} & \textbf{MMLU} & \textbf{Macro~Avg.} \\
\midrule
COT            & \textbf{0.831} & 0.030 & 0.414 & 0.187 & 0.610 & 0.578 & 0.675 & \cellcolor{gray!10}0.416 \\
TOT            & 0.810 & \underline{0.100} & 0.155 & \textbf{0.360} & 0.430 & \textbf{0.797} & \underline{0.735} & \cellcolor{gray!10}0.352 \\
Analogical & \underline{0.825} & 0.060 & 0.324 & 0.197 & \textbf{0.660} & 0.729 & 0.721 & \cellcolor{gray!10}0.433 \\
Self-Refine    & 0.716 & 0.030 & 0.213 & 0.167 & \underline{0.650} & \underline{0.796} & 0.543 & \cellcolor{gray!10}0.372 \\
SPP            & 0.823 & \textbf{0.160} & \textbf{0.536} & \underline{0.217} & 0.540 & 0.684 & 0.689 & \cellcolor{gray!10}\textbf{0.469} \\
STEP-BACK      & 0.817 & 0.010 & \textbf{0.536} & 0.190 & 0.570 & 0.642 & \textbf{0.788} & \cellcolor{gray!10}\underline{0.452} \\
SimTom         & 0.586 & 0.040 & 0.240 & 0.177 & 0.460 & 0.599 & 0.503 & \cellcolor{gray!10}0.315 \\
\textbf{MRP (our)} & 0.781 & 0.050 & 0.346 & 0.187 & 0.600 & 0.759 & 0.722 & \cellcolor{gray!10}0.433 \\
\bottomrule
\end{tabularx}
}
\end{table*}

\subsection{Main Results}

\paragraph{Meta-Reasoning Prompting performs best on comprehensive tasks}

\begin{figure}[t]
    \centering
    \includegraphics[width=0.85\columnwidth]{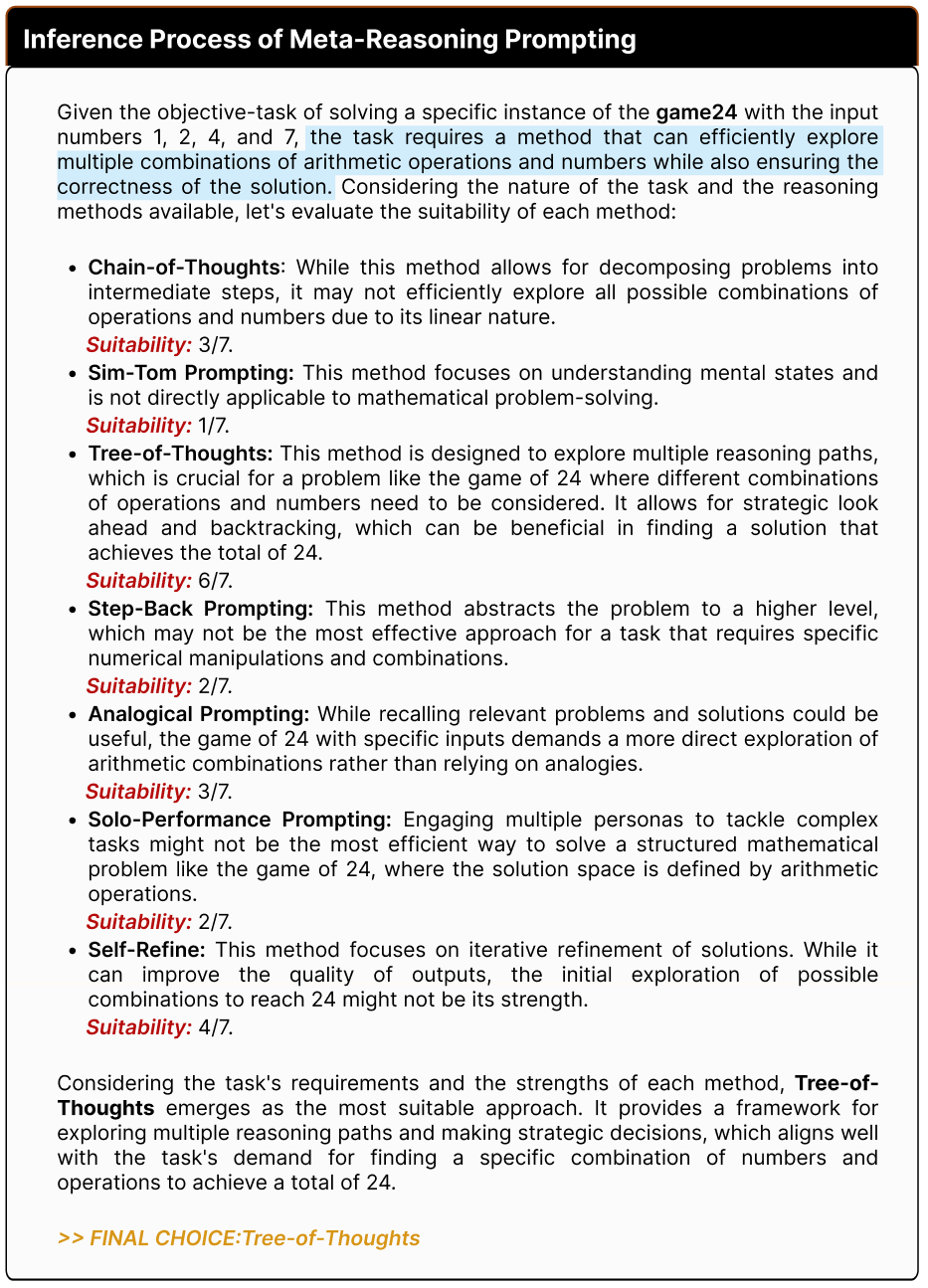}
    \caption{The inference process of large language models (LLMs) under meta-reasoning prompting.}
    \label{Inference Process of MRP}
\end{figure}

As shown in table \ref{tab:benchmarking_gpt4}, MRP consistently exhibits robust performance across multiple benchmarks. MRP achieves the \textbf{second-best} in \textbf{4 of 7} tasks, including \textbf{Gameof24}, \textbf{TriviaQA}, \textbf{BigToM} and \textbf{Code}. This impressive performance across a wide range of tasks demonstrates MRP's ability to effectively select and apply appropriate reasoning methods tailored to the specific requirements of each task. In terms of overall performance, MRP attains the highest across the 7 tasks, with an average of $0.772$. In contrast, although TOT excels in certain tasks such as GSM8K and Gameof24, it performs less impressively in others. We observe noticeable performance gaps compared with MRP in tasks such as BigToM (\textbf{0.43 VS 0.57}) and Code (\textbf{0.765 VS 0.867}). This consistent excellence across all benchmarks underscores MRP's advantages, demonstrating its ability to maintain impressive performance across diverse task domains (as shown in figure~\ref{Inference Process of MRP}).

\paragraph{Meta-reasoning capability is influenced by the base model capability}

As illustrated in table \ref{tab:benchmarking_gpt35}, while the performance with GPT-4 is satisfactory, the experimental results with GPT-3.5 indicate that the effectiveness of MRP is suboptimal. Error analysis revealed the main issues: \textit{Scoring Error}, \textit{Self-opinion}, \textit{Factual Error}, and \textit{Reasoning Error}. This indicates that when the model's capabilities are limited, it cannot have sufficient awareness of its own reasoning abilities and the meta-issues behind the reasoning problems. This performance drop also appears in other reasoning methods, which also indicates that the capability of meta-reasoning, like other reasoning abilities, improves as the model becomes more powerful.

\paragraph{Meta-Reasoning Prompting is less effective for simple tasks but significantly improved for more differentiated tasks}
\begin{wrapfigure}{r}{0.52\textwidth}
  \centering
\includegraphics[width=0.52\textwidth]{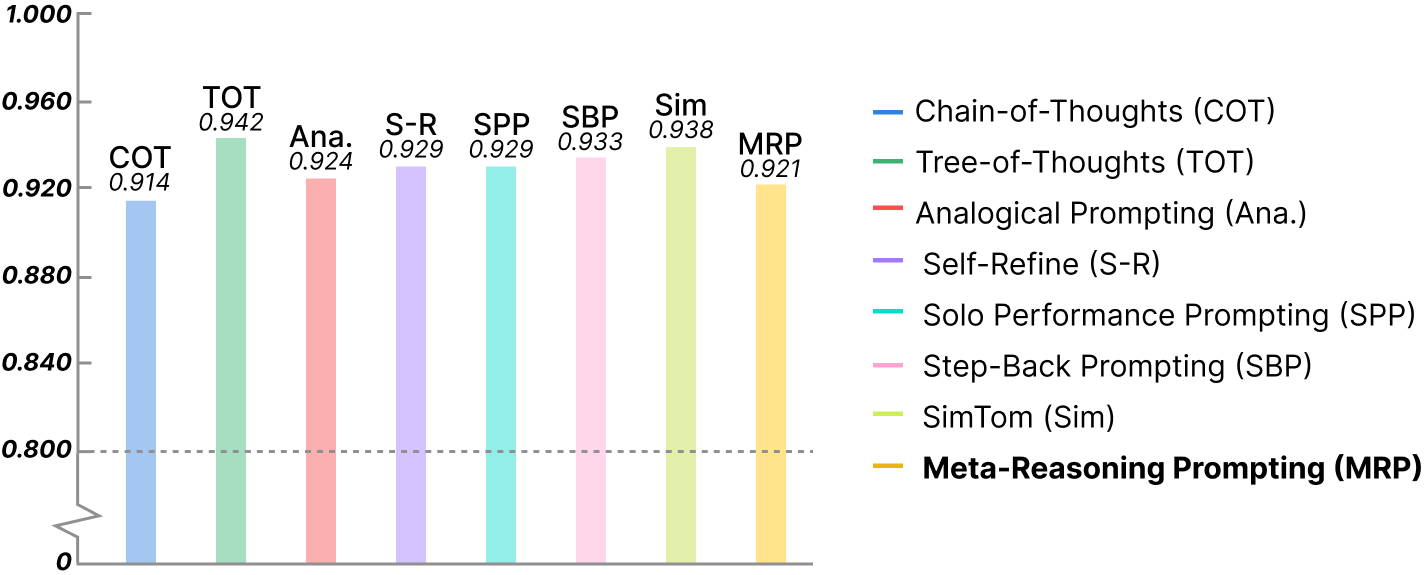}
\vspace{-0.1cm}
  \caption{Performance of methods on GSM8K benchmark}
  \vspace{-0.3cm}
  \label{fig:GSM8K}
\end{wrapfigure}
From the experimental results (see figure~\ref{fig:GSM8K}), it can be seen that MRP and other methods show equal competitiveness on GSM8K, the accuracy of all the reasoning methods is above 90\%, but the differentiation between the accuracy of each method is not very high, it can be seen that when the task is simpler, it is harder for MRP to reflect its own advantages, but MRP method is better than each method on the more difficult and comprehensive But the MRP method is significantly better than the other methods in the more difficult and comprehensive tasks.

\section{Related Works}

\subsection{Reasoning with LLMs}

Prompt-based reasoning methods have become a key technology for enhancing the capabilities of pretrained large language models (LLMs). The Chain-of-Thought (CoT) prompting~\cite{COT}, and its variants~\cite{LogiCOT,ThoT,BOT,COS,SC-COT}, such as Tree of thoughts (TOT)~\cite{TOT}, Graph of thoughts (GOT)~\cite{GOT}, enhances LLMs' ability to decompose complex tasks into smaller, manageable tasks, utilizing structured approaches to explore problem-solving pathways. Numerous studies have demonstrated the exceptional performance of prompt-based reasoning methods across various domains and benchmarks.~\cite{selfrefine,simtom,TSB,sahoo2024systematic,metaprompt} Some researchers have even employed analogical reasoning~\cite{yu2023thought,feng2024thought,yasunaga2023large}, enabling large models to generate similar questions based on user queries and subsequently summarize solutions based on the answers to these questions. While independent reasoning methods have been proven to improve LLM performance from different perspectives, they still fail to meet integrated problems.

There are also some methods to enhance LLM reasoning through ensemble mechanisms or tuning. X-of-Thoughts improves the success rate of LLM on arithmetic problems by integrating three methods~\cite{XOT}. It proposes a trial-and-error iterative mechanism that allows LLM to autonomously repeat attempts to find a final solution. Ni et al.blending off-the-shelf benchmarks to create a comprehensive, integrated LLM assessment~\cite{ni2024mixeval}.
Mixtural-Of-Prompts (MoP) dynamically manage and optimize prompt tuning across heterogeneous tasks and data distributions, significantly reducing perplexity and mitigating interference in multi-task scenarios~\cite{dun2023sweeping}. Some researchers fine-tune smaller models with a well-prepared dataset inspired by preference learning to achieve reasoning power comparable to a larger model~\cite{yuan2024advancing,sukhbaatar2024branch,MOI}. They present problem-method coupled datasets and show how to improve the model's grasp of inference skills at the data level.
However, there is still a lack of research to explore the meta-reasoning ability of LLMs to choose reasoning methods.

\subsection{Meta Reasoning}
Meta-reasoning is a crucial cognitive process in human intelligence, involving the recognition and interpretation of reasoning to select optimal methods based on past experiences~\cite{griffiths2019doing}. In artificial intelligence, it refers to efficiently deploying computational resources for informed decision-making in specific situations~\cite{cox2011metareasoning,cox2011metareasoningbook}. Recently, some works develop routing or buffer systems to improve performance, using supervised learning algorithms~\cite{shnitzer2023large}, reward model-based techniques, and other methods~\cite{hari2023tryage,lu2024blending, wang2024mixture}. Hu et al. created a benchmark to evaluate these methods' effectiveness~\cite{hu2024mars}. Zeng et al. noted the neglect of meta-reasoning in independent LLMs and proposed a benchmark to evaluate reasoning rationality~\cite{zeng2024mrgsm8k}. In \cite{yang2024buffer}, the authors introduce a meta-buffer to store a series of high-level thoughts distilled from problem-solving processes across various tasks. This approach aligns with the inherent logic of meta reasoning. However, MRP achieves simple and efficient meta-cognitive effects by directly unleashing the meta reasoning capabilities of LLM through prompts, without introducing complicated mechanisms.

\section{Conclusions and Outlook}
This paper introduces Meta-Reasoning Prompting (MRP), a novel and efficient approach inspired by human meta-reasoning, designed to enhance the adaptability and efficiency of large language models (LLMs). By dynamically selecting and applying the most suitable reasoning method for each task, MRP enables LLMs to optimize performance across diverse problem domains, achieving near state-of-the-art results in comprehensive benchmarks.

Our experiments demonstrate that MRP significantly improves LLMs' ability to handle tasks requiring a blend of different reasoning strategies, particularly in larger models like GPT-4. This dynamic adaptability highlights MRP's potential to address the limitations of traditional reasoning techniques, offering a more flexible and effective solution for varied and complex tasks.

Looking ahead, future research could explore the integration of MRP into training datasets to further enhance LLMs' general reasoning abilities. Additionally, combining MRP with other advanced reasoning techniques could yield further improvements in model performance. Overall, MRP represents a significant step forward in developing more intelligent, efficient, and adaptable AI systems, capable of meeting the diverse demands of real-world problem-solving.

\section{Limitations}
Our study investigates the meta-reasoning mechanisms of LLMs by dynamically selecting suitable methods to enhance their performance across various reasoning tasks without introducing new knowledge or training efforts. Currently, Meta-Reasoning Prompting (MRP) selects the highest-scoring method for each task. However, drawing from human cognitive processes, tackling complex problems often involves combining multiple reasoning methods. Future research will explore mechanisms such as Top-Probability (Top-P) or Top-K to allow models to ensemble relevant methods, potentially achieving better performance.

Our experimental results indicate that the meta-reasoning ability of LLMs is influenced by the capabilities of the models themselves. For instance, GPT-4's Meta-Reasoning Prompting shows significantly greater improvement compared to GPT-3.5, which aligns with our expectations. Nonetheless, we can further enhance the smaller model's meta-reasoning capabilities through instruction tuning in future works.

Due to space constraints and limited resources, our experiments primarily tested the most representative LLMs (GPT-4 and GPT-3.5). We did not fully cover the performance of other open-source or closed-source models. However, we believe that the experimental results on these representative LLMs provide sufficient insights and implications.

\bibliographystyle{plainnat}
\bibliography{custom}

\appendix

\section{Implementation Details}
\subsection{Dataset Details}
\label{sec:appendix Dataset Details}

Table~\ref{Dataset Split and Number of Examples} shows the split and number of examples used for evaluations in GSM8K, Game of 24, Trivia Creative Writing, HotpotQA, BigTOM, Code Readability and MMLU. The dataset sizes of GSM8K, Gameof24, Trivia Creative Writing are consistent with the size used in the references. To control cost, we randomly tested 100-300 sample of data from HotpotQA, BigTOM, and Code Readability and MMLU. Despite of the economic consideration, we found that on this data scale, MRP has achieved significant results.

\begin{table*}[ht]
    \centering
    \caption{Dataset Split and Number of Examples}
    \begin{tabular}{cccc}
        \toprule
        Domain & Dataset & Number of Examples \\
        \midrule
        Arithmetic Reasoning & GSM8K & 1319 \\
        Complex Mathematical Reasoning & Game of 24  & 100 \\
        Creative Writing & Trivia Creative Writing  & 100 \\
        Multi-hop Reasoning & HotpotQA  & 300 \\
        Social Reasoning & BigTOM  & 100 \\
        Computer Code & Code Readability  & 300 \\
        STEM & MMLU  & 151 \\
        \bottomrule
        \label{Dataset Split and Number of Examples}
    \end{tabular}
\end{table*}

\subsection{Source Prompts of the Reasoning Methods Used in This Paper}
\label{sec:appendix method prompt}

\begin{figure}[t]
    \centering
    \includegraphics[width=0.65\columnwidth]{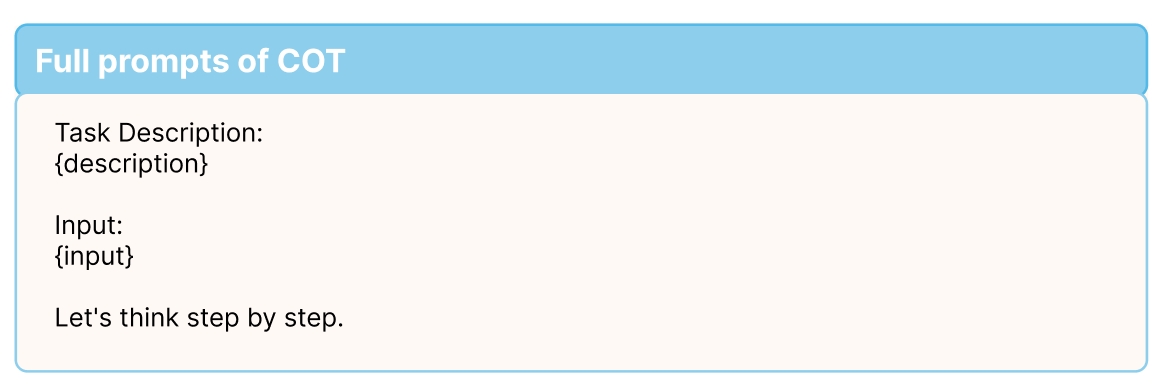}
    \caption{Prompt of COT}
    \label{Prompt of COT}
\end{figure}

\begin{figure}[t]
    \centering
    \includegraphics[width=0.65\columnwidth]{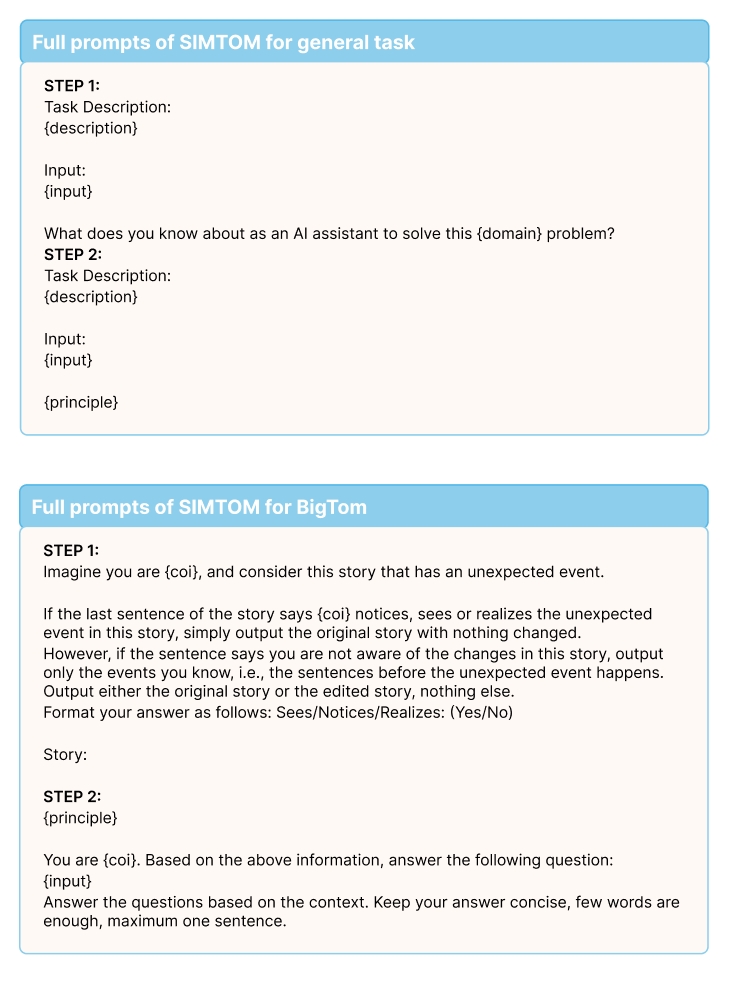}
    \caption{Prompt of Sim-Tom Prompting}
    \label{Prompt of Sim-Tom Prompting}
\end{figure}

\begin{figure}[t]
    \centering
    \includegraphics[width=0.65\columnwidth]{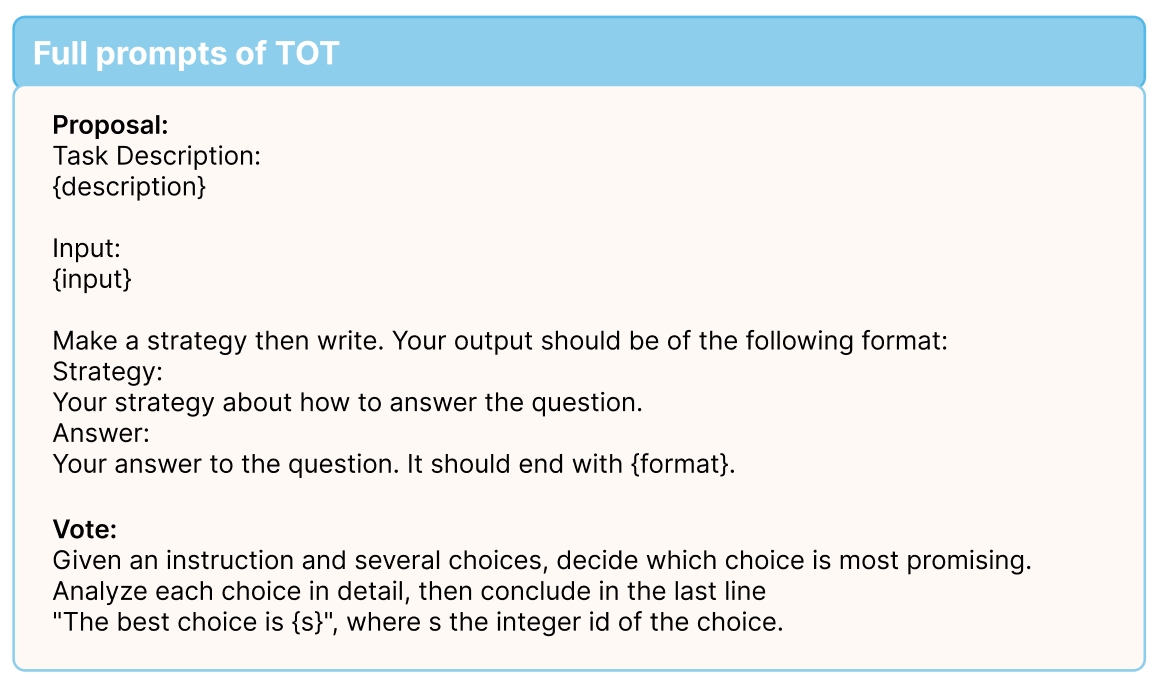}
    \caption{Prompt of TOT}
    \label{Prompt of TOT}
\end{figure}

\begin{figure}[t]
    \centering
    \includegraphics[width=0.65\columnwidth]{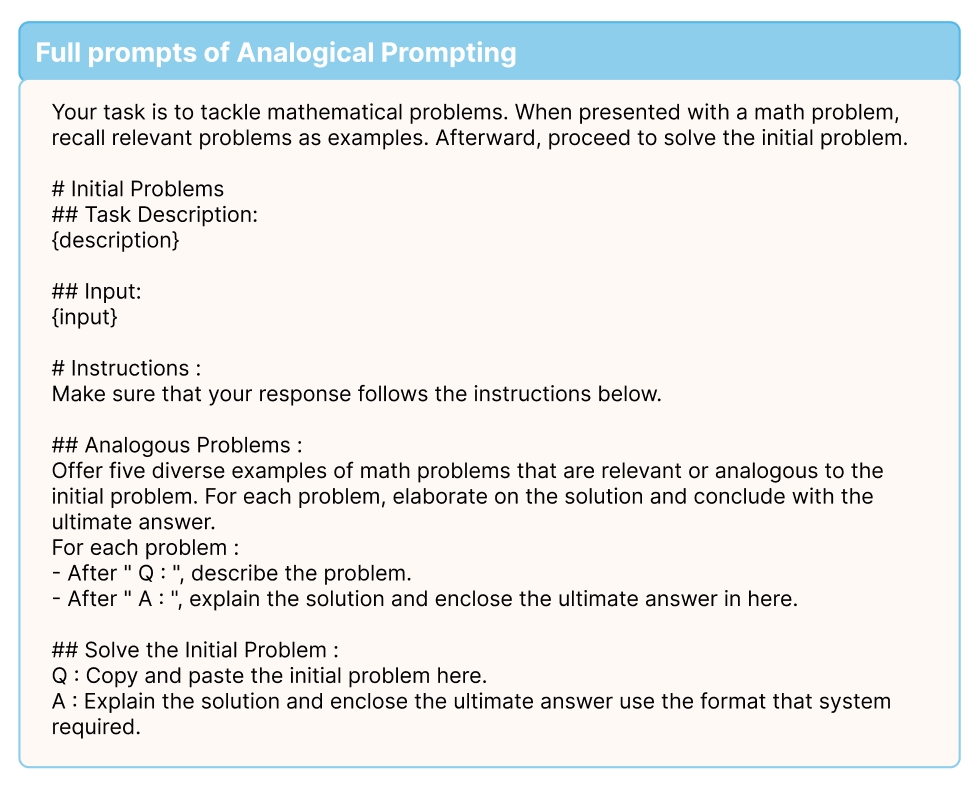}
    \caption{Prompt of Analogical Prompting}
    \label{Prompt of Analogical Prompting}
\end{figure}

\begin{figure}[t]
    \centering
    \includegraphics[width=0.65\columnwidth]{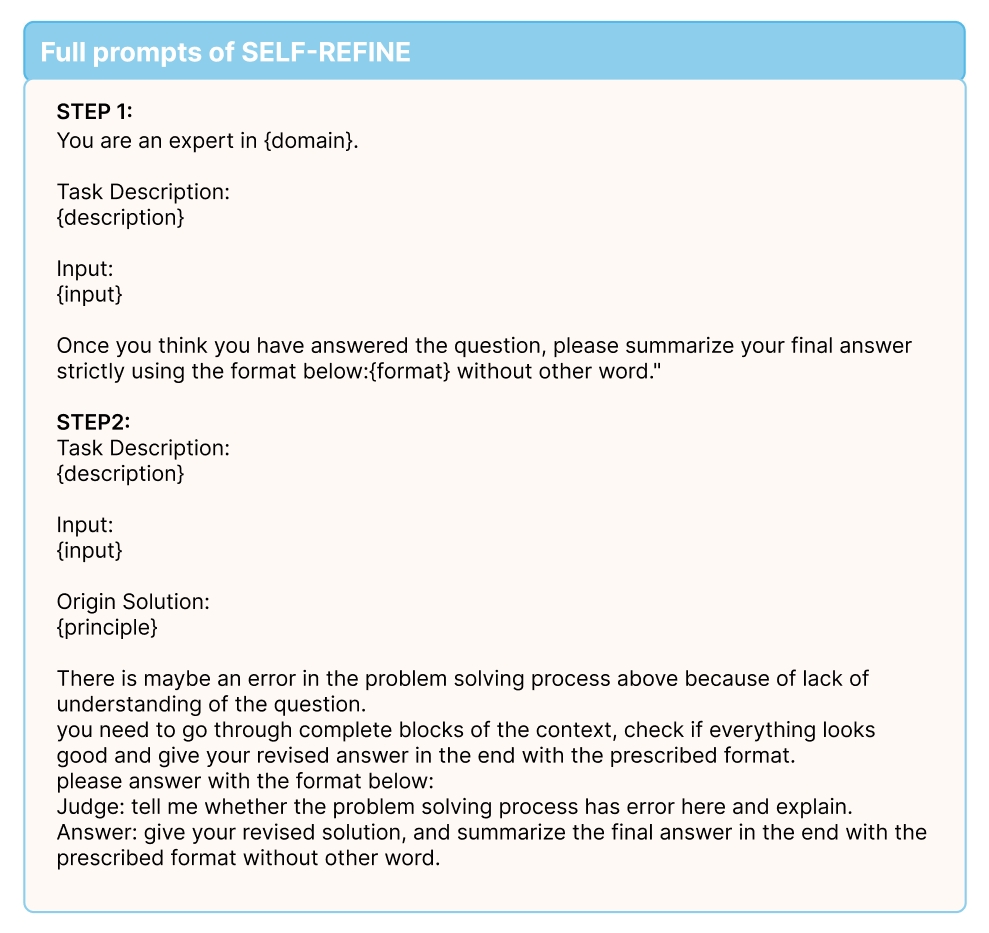}
    \caption{Prompt of SELF-REFINE}
    \label{Prompt of SELF-REFINE}
\end{figure}

\begin{figure}[t]
    \centering
    \includegraphics[width=0.65\columnwidth]{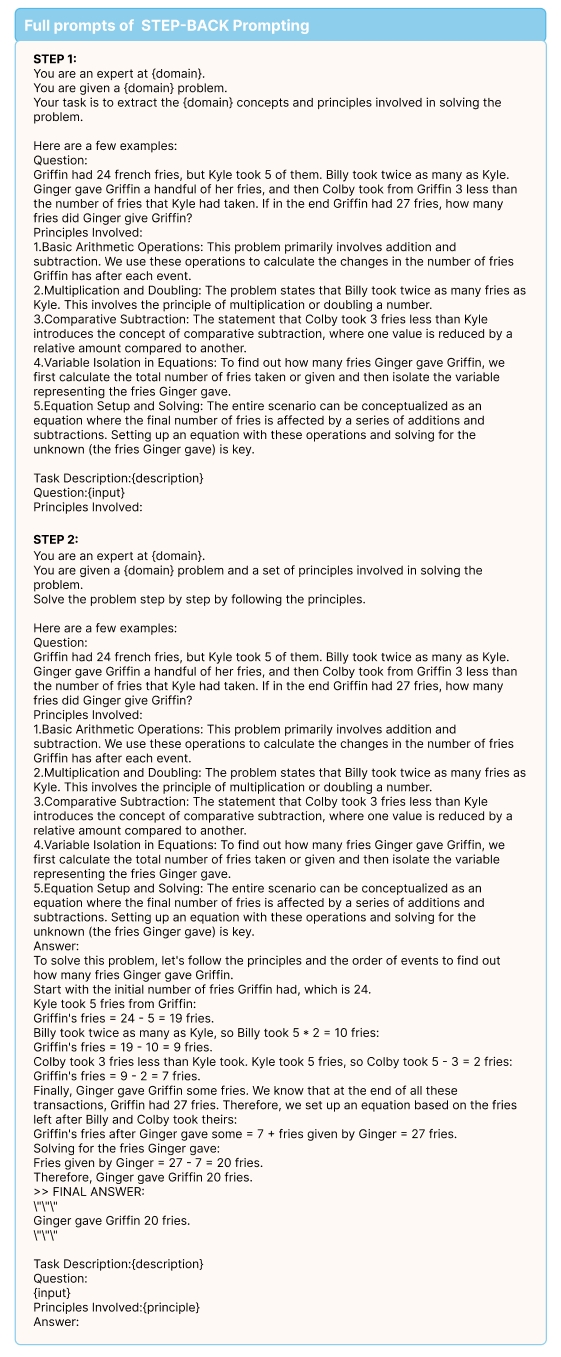}
    \caption{Prompt of STEP-BACK Prompting}
    \label{Prompt of STEP-BACK Prompting}
\end{figure}

\begin{figure}[t]
    \centering
    \includegraphics[width=0.65\columnwidth]{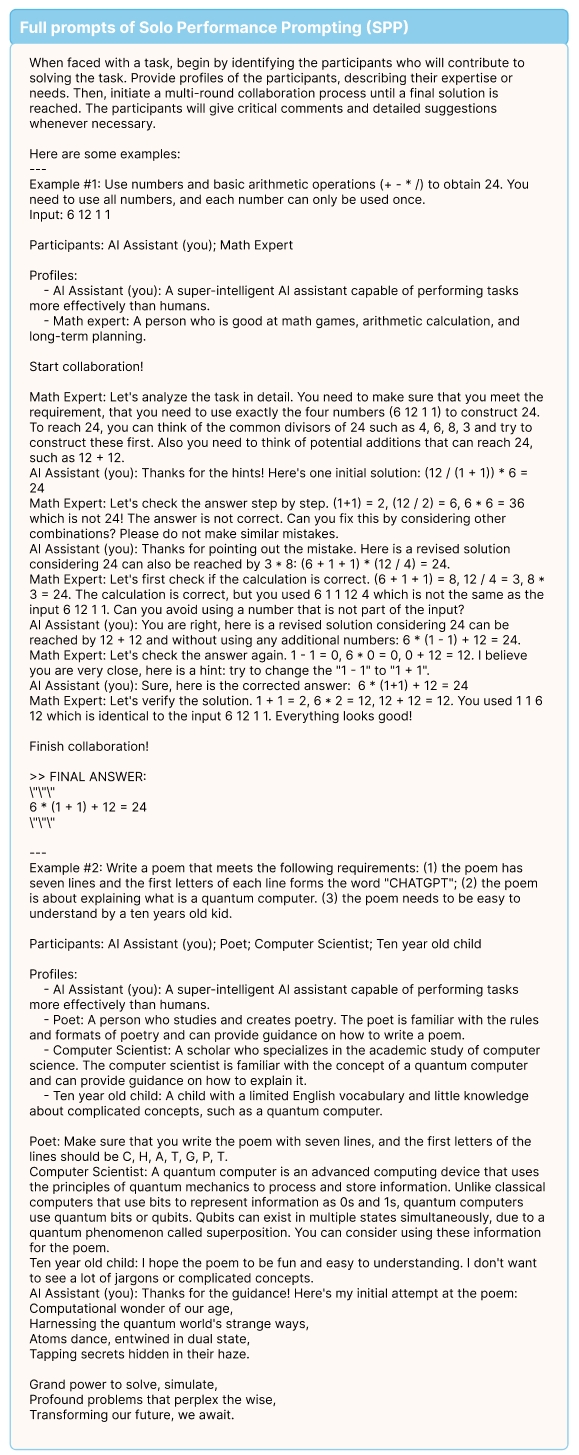}
    \caption{Prompt of SPP Prompting}
    \label{Part1 of SPP Prompt}
\end{figure}

\begin{figure}[t]
    \centering
    \includegraphics[width=0.65\columnwidth]{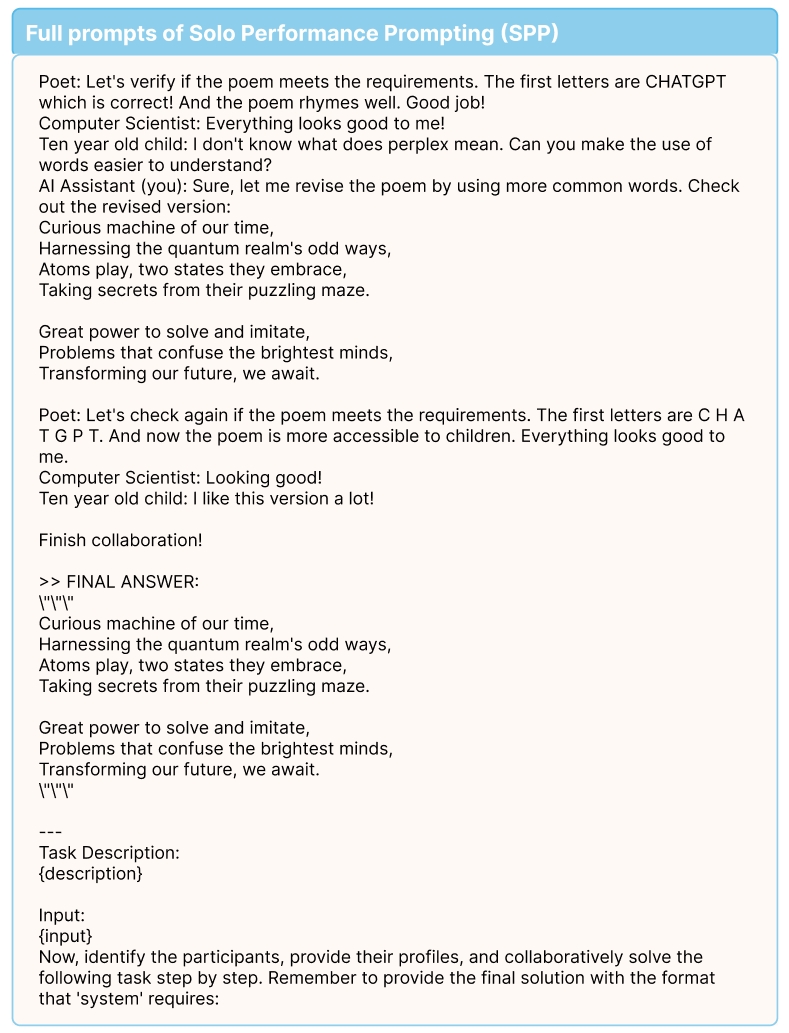}
    \caption{Prompt of SPP Prompting}
    \label{Part2 of SPP Prompt}
\end{figure}
\end{document}

%% file: settings.tex
\usepackage{multirow}
\usepackage{amsmath}
\usepackage{capt-of}
\usepackage{tabularx}
\usepackage{epsfig}
\usepackage{amssymb}
\usepackage{amsfonts}
\usepackage{booktabs}
\usepackage{scalerel}
\usepackage[inline]{enumitem}
\usepackage{listings}
\usepackage{varwidth}
\usepackage[export]{adjustbox}
\usepackage{tikz}
\usetikzlibrary{tikzmark}

\usepackage{stmaryrd}
\usepackage{bbm}
\usepackage{wrapfig}
\usepackage{pifont}
\usepackage[noabbrev]{cleveref}

\definecolor{deepblue}{rgb}{0,0,0.5}
\definecolor{officeblue}{RGB}{0,102,204}
\definecolor{deepred}{rgb}{0.6,0,0}
\definecolor{deepgreen}{rgb}{0,0.5,0}
\definecolor{mybrickred}{RGB}{182,50,28}

\definecolor{fillcolor}{RGB}{216,217,252}

\usepackage{etoolbox}
\usepackage{framed}

\newif\ifxetexorluatex
\ifxetex
  \xetexorluatextrue
\else
  \ifluatex
    \xetexorluatextrue
  \else
    \xetexorluatexfalse
  \fi
\fi
\newcommand*\quotesize{60} 
\newcommand*{\openquote}
   {\tikz[remember picture,overlay,xshift=-4ex,yshift=-2.5ex]
   \node (OQ) {\fontsize{\quotesize}{\quotesize}\selectfont``};\kern0pt}

\newcommand*{\closequote}[1]
  {\tikz[remember picture,overlay,xshift=4ex,yshift={#1}]
   \node (CQ) {\fontsize{\quotesize}{\quotesize}\selectfont''};}

\colorlet{shadecolor}{white}

\newcommand*\shadedauthorformat{\emph} 

\newcommand*\authoralign[1]{%
  \if#1l
    \def\authorfill{}\def\quotefill{\hfill}
  \else
    \if#1r
      \def\authorfill{\hfill}\def\quotefill{}
    \else
      \if#1c
        \gdef\authorfill{\hfill}\def\quotefill{\hfill}
      \else\typeout{Invalid option}
      \fi
    \fi
  \fi}
{\authoralign{#1}
\ifblank{#2}
   {\def\shadequoteauthor{}\def\yshift{-2ex}\def\quotefill{\hfill}}
   {\def\shadequoteauthor{\par\authorfill\shadedauthorformat{#2}}\def\yshift{2ex}}
\begin{snugshade}\begin{quote}\openquote}
{\shadequoteauthor\quotefill\closequote{\yshift}\end{quote}\end{snugshade}}

\newfloat{Code}{htbp}{Code}

%% file: math_commands.tex
\usepackage{amsmath,amsfonts,bm}

\def\eqref#1{equation~(\ref{#1})}

\def\1{\bm{1}}

\DeclareMathAlphabet{\mathsfit}{\encodingdefault}{\sfdefault}{m}{sl}
\SetMathAlphabet{\mathsfit}{bold}{\encodingdefault}{\sfdefault}{bx}{n}

